\documentclass[letterpaper, 10 pt, conference]{ieeeconf}  % Comment this line out
\IEEEoverridecommandlockouts                              % This command is only
\usepackage[backend=bibtex, style=ieee, natbib=true, isbn=false]{biblatex}
\AtEveryBibitem{\clearlist{language}}

\usepackage{graphicx} % for pdf, bitmapped graphics files
\usepackage{amsmath, amssymb, textcomp, gensymb, bm, mathtools}
\usepackage{color}
\newcommand{\SE}{\mathrm{SE}}
\newcommand{\se}{\mathfrak{se}}
\newcommand{\SO}{\mathrm{SO}}
\newcommand{\so}{\mathfrak{so}}
\newcommand{\UAV}{\mathrm{G}}
\newcommand{\uav}{\mathfrak{g}}
\newcommand{\g}{\mathfrak{g}}

\newcommand{\inv}{^{\text{-}1}}

\newcommand*{\Real}[1]{\ensuremath{\mathbb{R}^{#1}}}

\newcommand\norm[1]{\ensuremath{\left\Vert #1 \right\Vert}}

\newcommand\projsym[1]{\ensuremath{\mathbb{P}_s\left( #1 \right)}}

\DeclareMathOperator*{\argmin}{arg\,min}

\DeclareMathOperator{\blkdiag}{blkdiag}

\DeclareMathOperator{\ad}{ad}
\DeclareMathOperator\Hess{Hess}
\bmdefine{\der}{\mathrm{d}}
\DeclareMathOperator{\tr}{tr}

\newcommand{\gravity}{g}
\newcommand{\wbias}{\theta}
\newcommand{\abias}{\phi}
\newcommand{\nvehicles}{n}
\newcommand{\nlandmarks}{n_L}
\newcommand{\DQD}{M}
\newcommand{\connection}{\Gamma}
\newcommand{\Rab}{{}_{\alpha} R_\beta}

\title{\LARGE \bf Inertial Collaborative Localisation for Autonomous Vehicles using a Minimum Energy Filter}

\author{Jack Henderson, Mohammad Zamani, Robert Mahony, and Jochen Trumpf% <-this % stops a space
\thanks{
		Jack Henderson, Robert Mahony, and Jochen Trumpf are with Systems Theory and Robotics at the Australian National University, \texttt{firstname.lastname@anu.edu.au}.	Mohammad Zamani is with the Land Division, Defence Science and Technology Group, Australia, \texttt{mohammad.zamani@dst.defence.gov.au}}%
\thanks{This research is supported by the Commonwealth of Australia as represented by the Defence Science and Technology Group of the Department of Defence and by the Australian Research Council Discovery Project DP190103615: ``Control of Network Systems with Signed Dynamical Interconnections''}%
}

\begin{document}
\maketitle
\thispagestyle{empty}
\pagestyle{empty}

\begin{abstract}
    Collaborative Localisation has been studied extensively in recent years as a way to improve pose estimation of unmanned aerial vehicles in challenging environments.
    However little attention has been paid toward advancing the underlying filter design beyond standard Extended Kalman Filter-based approaches. 
    In this paper, we detail a discrete-time collaborative localisation filter using the deterministic minimum-energy framework.
    The filter incorporates measurements from an inertial measurement unit and models the effects of sensor bias and gravitational acceleration. 
    We present a simulation based on real-world vehicle trajectories and IMU data that demonstrates how collaborative localisation can improve performance over single-vehicle methods.
\end{abstract}
    
\section{Introduction}

Accurate localisation, and specifically pose estimation, for mobile robotic vehicles remains a challenge in environments where sensor information is limited or degraded.
The traditional use of GNSS technology for localisation is unsuitable for indoor or underground environments where satellite signals cannot be received, or in cases where radio interference is present.
Where multiple vehicles are operating in the same environment, these vehicles can share information between neighbours and improve the accuracy of their estimated pose  through process known as collaborative localisation (CL) \cite{roumeliotisDistributedMultirobotLocalization2002}.
However, as information propagates through the network, one must be careful to account for the introduced correlations in the state space.
Incorrectly doing so can result in overly confident state estimates \cite{kiaCooperativeLocalizationMobile2016} and this presents one of the main challenges in collaborative localisation.

Roumeliotis and Bekey \cite{roumeliotisDistributedMultirobotLocalization2002} present an Extended Kalman Filter (EKF) which estimates the joint state of a network of ground-based robots. By tracking the joint covariance matrix of the network, they correctly account for cross-correlations between vehicles.
They also show how, instead of requiring a single state estimator to centrally compute the estimate for the entire network, an equivalent formulation can be constructed where each vehicle runs a local estimator.
Subsequent work has built on this theory, but the focus has been towards finding different approximations of the same EKF-based filter that reduce the communication complexity; for example, by using techniques such as Covariance Intersection \cite{carrillo-arceDecentralizedMultirobotCooperative2013, mokhtarzadehCooperativeInertialNavigation2014}, alternative communication strategies \cite{kiaCooperativeLocalizationMobile2016}, or by conservatively approximating cross-covariance terms with different information \cite{luftRecursiveDecentralizedLocalization2018}.

In contrast to this, there has been significant research in filtering theory, particularly for attitude estimation.
New filters such as the Multiplicative EKF \cite{markleyAttitudeErrorRepresentations2003, martinGeneralizedMultiplicativeExtended2010}, the complementary filter \cite{mahonyComplementaryFilterDesign2005,mahonyObserversKinematicSystems2013}, and the Invariant EKF \cite{bonnabelSymmetryPreservingObservers2008,bonnableInvariantExtendedKalman2009} show significant improvements in performance over the standard EKF.
More recently, minimum energy filtering \cite{mortensenMaximumlikelihoodRecursiveNonlinear1968, sacconSecondOrderOptimalMinimumEnergyFilters2016,zamaniDiscreteUpdatePose2019} has been proposed as an alternative to conventional stochastic filtering methods.
Apart from the authors own work \cite{hendersonMinimumEnergyFilter2020IFAC,zamaniCollaborativePoseFiltering2019}, we are unaware of other formulations that utilise these modern filtering techniques in the context of the collaborative localisation problem.

In this paper, we build on our two previous works on this topic to create a distributed collaborative localisation algorithm capable of being implemented on a real network of mobile robots.
We extend the work from \cite{hendersonMinimumEnergyFilter2020CDC} to account for biases in the inertial measurement unit, as well as the additional effects of gravity on the measurements.
We also draw from our work in \cite{hendersonMinimumEnergyFilter2020IFAC} to extend the single vehicle filter into a multi-vehicle collaborative filter.
Our simulations are based on data from real-world experiments and demonstrate that well posed collaborative filtering leads to a significant performance improvement over non-collaborative filter implementation. 
To authors knowledge, the present contribution is the first to provide a 15 degrees-of-freedom (DoF) multi-robot algorithm capable of dealing with the biases and multi-rate sampling associated with implementation on real-world systems.

The remainder of the paper is structured as follows;
In Section \ref{sec:preliminaries} we provide a brief overview of the notation used throughout the paper, and introduce the Lie group used for the state representation.
We formally define the state estimation problem in Section \ref{sec:problem-formulation}, including the system kinematics and the measurement models used.
In Section \ref{sec:joint_filter}, we describe a continuous-time minimum energy filter to estimate the joint state of the network of vehicles.
We then detail in Section \ref{sec:discrete_time} how the continuous time filter can be discretised, given that sensors produce measurements at different time intervals.
Section \ref{sec:distributed-filter} then shows that the equations of the centralised joint filter can be decoupled and computed locally on each vehicle in the network and we discuss the communication requirements this introduces.
In Section \ref{sec:simulation}, we demonstrate the capabilities of the proposed filter in a simulation which uses real-world vehicle trajectories and inertial data, combined with synthetically generated landmark and inter-vehicle measurements.
We then summarise our findings in Section \ref{sec:conclusion}.
\section{Preliminaries}
\label{sec:preliminaries}
In this section, we introduce some of the notation and concepts that will be used in the remainder of the paper.

\subsection{Notation}
A brief explanation of the symbols used in this paper is given below. A more comprehensive explanation of some of these symbols can be found in \cite{sacconSecondOrderOptimalMinimumEnergyFilters2016}.
\begin{itemize}
    \item $\projsym{X} = \frac{1}{2}(X + X^\top)$ is the symmetric projection
    \item $\blkdiag$ is the block-diagonal matrix constructor
    \item $(.)_\times : \Real{3} \rightarrow \so(3)$ is the skew-symmetric operator such that the cross product $a \times b = (a_\times) b$
    \item $[\Psi, \Phi]$ is the Lie bracket operator
    \item $T_e L_X \circ \Psi$ is the tangent map at $e$ of the left translation of $\Psi$, denoted by the shorthand $X \Psi$
    \item $\der f(X) \circ X\Psi$ is the directional derivative of $f$, evaluated at $X$, in direction $X \Psi$
    \item $\Hess$ is the Hessian operator

    \item $\mathfrak{g}^*$ is the dual of the Lie algebra $\mathfrak{g}$
    \item $\ad_\Psi$ is the adjoint representation of the Lie algebra
    \item $(.)^{T_X G}$ is the exponential functor
\end{itemize}

\subsection{Vehicle state representation}
The extended special Euclidean group, $\SE_2(3)$, introduced by \cite{barrauEKFSLAMAlgorithmConsistency2016} provides a convenient way to represent a vehicle's rotation, $R$, position, $x$, and velocity, $v$, as a matrix Lie group. The matrix representation of the group is
\begin{align*}
   P &= \begin{bmatrix}
      R & x & v \\ 0 & 1 & 0 \\ 0 & 0 & 1 
  \end{bmatrix} \in \SE_2(3).
\end{align*}

We extend $\SE_2(3)$ with two additional vectors, $\wbias, \abias \in \Real{3}$, to create the direct product group $\UAV$, which is used to represent the full 15-degrees-of-freedom (DoF) state of a single vehicle, namely the pose, velocity, and IMU sensor biases;
\begin{align*}
   \UAV &\coloneqq \SE_2(3) \times \Real{3} \times \Real{3},
   \\
   X &= (P, \wbias, \abias) \in \UAV.
\end{align*}
For $X, Y \in \UAV$, the group product operation is
\begin{align*}
   X \cdot Y &\coloneqq  (P_X P_Y, \wbias_X + \wbias_Y, \abias_X + \abias_Y)
\end{align*}
and the lie algebra, $\uav$, has the structure
\begin{gather*}
   \Psi = (\Psi_P, \Psi_{\wbias}, \Psi_{\abias}) \in \uav
   \intertext{where}
   \Psi_P = \begin{bmatrix} (\Psi_R)_\times & \Psi_x & \Psi_v\\ 0 & 0 & 0 \\ 0 & 0 & 0 \end{bmatrix}\in \se_2(3),
   \\
   \Psi_R, \Psi_x, \Psi_v, \Psi_{\wbias}, \Psi_{\abias} \in \Real{3}.
\end{gather*}
This yields the following properties for $X \in \UAV$ and $\Psi, \Phi \in \uav$;
\begin{align}
   \nonumber
   X\inv &= (P\inv, -{\wbias}, -{\abias}),
    \\
    \nonumber
    [\Psi, \Phi] &= ([P_\Psi, P_\Phi], 0, 0),
    \\
    T_e L_X (\Psi) &= X\Psi = (P \Psi_P, \Psi_{\wbias}, \Psi_{\abias}).
    \label{eq:left_translation}
    % \\
    % [P_\Psi, P_\Phi] &= (\Psi_{R\times}\Phi_{R\times} - \Phi_{R\times}\Psi_{R\times}, \Psi_{R\times}\Phi_v - \Phi_{R\times}\Psi_v,
    % \\ &\qquad \Psi_{R\times}\Phi_x - \Phi_{R\times}\Psi_x ).
\end{align}

Further to this, the state of multiple vehicles in the network can be represented with another direct product group, $G(\nvehicles)$, which is simply defined as
\begin{align*}
   G(\nvehicles) := \underbrace{G\times \ldots \times G}_\nvehicles.
\end{align*}

\subsection{Wedge and Vee operator}
The wedge and vee operators transform between the Lie algebra and the vector representation as follows;
\begin{align*}
\vee &: \g \rightarrow \Real{15},
&
\wedge &: \Real{15} \rightarrow \g,
\\
\Psi^\vee &\coloneqq \begin{bsmallmatrix}
   \Psi_R \\ \Psi_x \\ \Psi_v \\ \Psi_{\wbias} \\ \Psi_{\abias}
\end{bsmallmatrix},
&
(\Psi^\vee)^\wedge &\coloneqq \Psi.
\end{align*}
The notation allows us to represent operators on the Lie algebra as matrix operators on the vector representation of the Lie algebra.
For example, consider the adjoint representation of $\g$, which is equivalent to the Lie bracket;
\begin{gather*}
   \ad(\Psi, \Phi) = \ad_\Psi \circ \Phi = [\Psi, \Phi].
\end{gather*}
We can define a new matrix operator $\check{\ad} \in \Real{15 \times 15}$, which is related to the original adjoint operator by
\begin{gather*}
   \check{\ad}_\Psi \Phi^\vee = (\ad_\Psi \circ \Phi)^\vee.
\end{gather*}
The matrix representation is then
\begin{gather*}
\check{\ad}_\Psi \coloneqq \begin{bmatrix}
   \Psi_{R\times} & 0 & 0 & 0 & 0 \\
   \Psi_{x\times} & \Psi_{R\times} & 0 & 0 & 0 \\
   \Psi_{v\times} & 0 &  \Psi_{R\times} & 0 & 0 \\
   0 & 0 & 0 & 0 & 0  \\
   0 & 0 & 0 & 0 & 0 
\end{bmatrix}.
\end{gather*}

\subsection{Connection Function}
The left-invariant affine connection on $G$ is characterised by the connection function $\connection: \g \times \g \rightarrow \g$.
While there are a number of possible choices for the connection function, in this paper we select the Cartan (0)-Connection which is defined as
\begin{align}
\connection(\Psi, \Phi) &\coloneqq \frac{1}{2} [\Psi, \Phi] = \frac{1}{2} \ad_\Psi(\Phi).
\label{eq:def_connection}
\end{align}
For the chosen connection, the torsion tensor is zero, i.e. $T(\Psi, \Phi) = 0$.

\section{Problem Formulation}
\label{sec:problem-formulation}
Consider the problem of determining the pose, both position and orientation, of a network of $\nvehicles$ vehicles which are free to move in 3-dimensional space.
The set of vehicles in the network is defined as $\bm{V} \coloneqq \{ 1, \ldots, \nvehicles \}$, and the indices $\alpha$ and $\beta \in \bm{V}$ are used throughout this paper to refer to particular vehicles in the network.
These vehicles could be unmanned aerial vehicles (UAV), unmanned ground vehicles (UGV), or a combination of both.

The vehicles move freely through a known environment where there are a set of $\nlandmarks$ fixed landmark points, indexed by the set $\bm{L} \coloneqq \{1, \ldots, \nlandmarks \}$. For a given landmark point, $i\in \bm{L}$, the known position of the landmark w.r.t. the inertial frame is $l_i \in \Real{3}$.

In terms of sensors, each vehicle is equipped with a strap-down inertial measurement unit (IMU), which measures the linear acceleration and angular velocity of the vehicle.
In addition to this, vehicles are also equipped with a sensor capable of measuring relative positions to the known landmarks in the environment, as well as relative positions to other vehicles in the network.

The task of collaborative localisation is to determine an estimate of the pose of each vehicle in the network at the current time, $t$.
Initially, we will pose this as a joint filtering problem, where the state of the entire network of vehicles is centrally estimated.
However, a centralised algorithm poses a number of challenges and imposes a single point of failure for the system.
In Section \ref{sec:distributed-filter}, we show how the same equations that implement the centralised filter can be decoupled so that each vehicle computes its own state estimate.

\subsection{Single Vehicle System Model}

Consider a rigid-body vehicle in free 3-D space.
The orientation, $R$, position, $x$, and linear velocity, $v$ of the body-fixed frame with respect to the inertial frame are all expressed in the coordinates of the inertial frame.
The kinematics of the vehicle is then modelled by
\begin{align*}
R &\in \SO(3),
&
x &\in \Real{3},
&
v &\in \Real{3},
\\
\dot{R} &= R \omega_\times,
&
\dot{x} &= v,
&
\dot{v} &= R a,
\end{align*}
where $\omega \in \Real{3}$ is the angular velocity, and $a \in \Real{3}$ is the linear acceleration of the body-fixed frame with respect to the inertial frame, expressed in the body-fixed frame.

Measurements from MEMS IMU sensors are commonly prone to time-varying biases \cite{tittertonStrapdownInertialNavigation2004}, which can introduce significant error if unaccounted for. To account for this, we model an offset of $\wbias$ to the angular velocity measurements and an offset of $\abias$ to the linear acceleration measurements. These biases vary slowly over time according to some unknown processes, $ \delta_{\wbias}$ and $ \delta_{\abias}$, weighted by $B_\wbias, B_\abias \in \Real{3 \times 3}$ respectively. This gives
\begin{align*}
\wbias &\in \Real{3},
&
\abias &\in \Real{3},
\\
\dot{\wbias} &= B_{\wbias} \delta_{\wbias},
&
\dot{\abias} &= B_{\abias} \delta_{\abias}.
\end{align*}

Combined, the state of a single vehicle, $\alpha \in \bm{V}$, is represented as
\begin{align*}
   X^\alpha &\coloneqq \left(
      \begin{bmatrix}
      R^\alpha & x^\alpha & v^\alpha \\ 0 & 1 & 0 \\ 0 & 0 & 1
      \end{bmatrix},
      \wbias^\alpha, \abias^\alpha
   \right) \in \UAV.
\end{align*}
Recalling \eqref{eq:left_translation}, the left-invariant kinematics are
\begin{gather*}
   \dot{X}^\alpha = X^\alpha \left(
       \begin{bmatrix}\omega^\alpha_\times & (R^\alpha)^\top v^\alpha & a^\alpha \\ 0&0&0 \\ 0&0&0 \end{bmatrix},
       B_{\wbias} \delta^\alpha_{\wbias},
       B_{\abias} \delta^\alpha_{\abias}
       \right),
\end{gather*}
where $X^\alpha(t_0) = X^\alpha_0$ is the initial state of the vehicle.

\subsection{IMU Sensor Model}
The vehicle, $\alpha$, is equipped with an IMU, which measures linear acceleration $u^\alpha_a$ and angular velocity, $u^\alpha_\omega$, in the body-fixed frame. The measurement vector $u^\alpha$ is then
\begin{align*}
    u^\alpha &\coloneqq \begin{bmatrix}u^\alpha_\omega \\ u^\alpha_a \end{bmatrix}
        = \begin{bmatrix} \omega^\alpha + \wbias^\alpha + B_\omega \delta^\alpha_\omega
            \\
            a^\alpha + \abias^\alpha + (R^\alpha)^\top \gravity +  B_a \delta^\alpha_a
        \end{bmatrix}
   \in \Real{6},
\end{align*}
which is composed of the true angular velocity, $\omega$, true acceleration, $a$, the time-varying biases, $\wbias$ and $\abias$, and $\delta^\alpha_\omega, \delta^\alpha_a \in \Real{3}$ which are unknown error signals weighted by $B_\omega, B_a \in \Real{3 \times 3}$. The accelerometer also measures acceleration due to gravity, $\gravity \approx 9.81 \cdot \text{e}_3$.

Substituting the measurement model into the vehicle kinematics gives
\begin{gather}
\dot{X}^\alpha = X^\alpha \left( \lambda^\alpha(X^\alpha, u^\alpha) + B^\alpha(\delta^\alpha )\right)
\intertext{where $\lambda^\alpha: \UAV \times \Real{6} \rightarrow \uav$ is given by}
\lambda^\alpha(X, u)^\vee \coloneqq \begin{bmatrix}
   u_\omega -\wbias \\
   R^\top v \\
   u_a - \abias - R^\top \gravity\\
   0 \\
   0
\end{bmatrix},
\label{eq:def_lambda}
\end{gather}
and the linear map $B^\alpha : \Real{12} \rightarrow \uav$ is given by
\begin{align}
B^\alpha(\delta)^\vee &\coloneqq 
\check{B}^\alpha
\begin{bmatrix}
   \delta_\omega \\ \delta_a \\ \delta_{\wbias} \\ \delta_{\abias}
\end{bmatrix}, 
&
\check{B}^\alpha &\coloneqq \begin{bsmallmatrix}
    -B_\omega & 0 & 0 & 0\\
    0 & 0 & 0 & 0 \\
    0 & - B_a & 0 & 0 \\
    0 & 0 & B_{\wbias} & 0 \\
    0 & 0 & 0 & B_{\abias} 
 \end{bsmallmatrix}.
\label{eq:def_B}
\end{align}

% For simplicity of notation, we will assume that the linear map $B^\alpha(\delta)$ is identical for all vehicles in the network, however it is straightforward to show otherwise.

\subsection{Joint System Model}
Consider the state of the entire network of vehicles, $X$, as an element of the $\UAV(n)$ group;
\begin{align*}
    X \coloneqq (X^1, \ldots, X^{\nvehicles}) \in \UAV(\nvehicles).
\end{align*}
The kinematics of the network can be modelled by 
\begin{align}
    \dot{X} &= X \left( \lambda(X, u) + B(\delta )\right)
\end{align}
where $\lambda$ and $B$ are a concatenation of the single-vehicle models described in \eqref{eq:def_lambda} and \eqref{eq:def_B} as follows;
\begin{gather}
    \begin{aligned}
    \lambda(X, u)^\vee &= \begin{bmatrix}
        \lambda^1(X^1, u^1)^\vee \\
        \vdots \\
        \lambda^n(X^{\nvehicles}, u^{\nvehicles})^\vee \\
    \end{bmatrix},
    & 
    B(\delta)^\vee &= \check{B} \begin{bmatrix}
        \delta^{1}\\
        \vdots \\
        \delta^{\nvehicles}
    \end{bmatrix},
\end{aligned}
\nonumber
\\
\check{B} = \blkdiag(\check{B}^1, \ldots, \check{B}^{\nvehicles}).
\label{eq:def_check_B}
\end{gather}

\subsection{Measurement Model}
In addition to the IMU, each vehicle is equipped with a sensor that measures relative translations to a number of fixed landmarks, $i \in \bm{L}$, in the environment.
The sensor measurement made by vehicle $\alpha$ of a single landmark $i$ is denoted as $y^\alpha_i$.
It is modelled as the true relative translation, $h$, corrupted by some unknown measurement error, $\epsilon^\alpha_i \in \Real{3}$;
\begin{align}
h(X, l) &= R^\top(l - x),
\\
y^\alpha_i(t) &= h(X^\alpha(t), l_i) + D \epsilon^\alpha_i(t),
\label{eq:def_landmark_measurement}
\end{align}
where $D \in \Real{3\times 3}$ is invertible.

We can model the inter-vehicle measurements in a similar way to the landmark measurements. The landmark, $l$, is substituted by a marker point $m_\beta \in \Real{3}$ located at a known fixed point in the body-fixed frame of vehicle $\beta \in \bm{V}$. A measurement made at time $t$ of the marker on vehicle $\beta$, received by vehicle $\alpha$  is denoted by
\begin{align*}
    y_\beta^\alpha(t) &= h(X^\alpha(t), X^\beta(t)) + D \epsilon_\beta^\alpha(t),
    \\
    h(X^\alpha, X^\beta) &= (R^\alpha)^\top (R^\beta m_\beta + x^\beta - x^\alpha).
\end{align*}

\section{The Joint Minimum Energy Filter}
\label{sec:joint_filter}
In this section we define the minimum energy filtering problem on the joint system and present the resulting filter equations for the second-order optimal minimum energy filter.

Initially, we  pose the filtering problem in continuous time and suppose that both the IMU measurements, $u$, and the sensors measurements, $y$, are continuous-time signals that are always present.
The corresponding error signals, $\delta$ and $\epsilon$, are then also continuous-time signals, and we assume them to the zero mean and square integrable.

By using this formulation, we can take advantage of the existing body of work in the literature on continuous-time minimum-energy filtering.
This will yield a state estimate in the form of a continuous-time differential equation.
Later in this paper, in Section \ref{sec:discrete_time}, we discuss a more realistic discrete model for sensor measurements and propose a way to adapt the continuous-time model to account for the differences.

\subsection{Minium Energy Filtering Problem Definition}
We will consider the following cost functional on the system;
\begin{multline}
    J_t(\delta_{[t_0, t]}, \epsilon_{[t_0, t]}, X_0)
    \coloneqq \frac{1}{2}J_0(X_0)
    + \frac{1}{2} \int_{t_0}^t
        \norm{\delta(\tau)}^2_W
        \\
        + \sum_{\mathclap{i\in \bm{L}, \alpha \in \bm{V}}} \norm{\epsilon_i^\alpha(\tau)}^2_{Q^\alpha_i}
        + \sum_{\mathclap{\alpha, \beta  \in \bm{V}}} \norm{\epsilon_\beta^\alpha(\tau)}^2_{Q^\alpha_\beta}
        \mathrm{d} \tau
        \label{eq:def_cost_functional}
 \end{multline}
 where $J_0: G \rightarrow \Real{}$ is some cost on the initial state, $X_0$, with a unique global minimum. $Q^\alpha_\beta, Q^\alpha_i \in \Real{3 \times 3}$ are symmetric positive definite matrices which can be used to weight the norms of the respective error signals. $W$ is also symmetric positive definite, defined by $W=\blkdiag(W^1, \ldots, W^n) \in \Real{12 \nvehicles \times 12 \nvehicles}$.

The optimal trajectory of the system, $X^*_{[t_0, t]}$, is defined as the trajectory which satisfies the kinematics of the system and the measurement model, and which minimises the cost functional.
We then define the estimate of the state of the system, $\hat{X}(t)$ as the terminal point of the optimal trajectory up to the current time, $t$.
In other words, $\hat{X}(t) \coloneqq X^*_{[t_0, t]}(t)$.

\subsection{The Second-Order Minimum Energy Filter}
\label{sec:minimum_energy_abstract}
The filtering problem described in Section \ref{sec:problem-formulation} and the accompanying cost functional in \eqref{eq:def_cost_functional} is a particular case of the filtering problem described in \cite{sacconSecondOrderOptimalMinimumEnergyFilters2016}. Given this, we adapt the solution from \cite{sacconSecondOrderOptimalMinimumEnergyFilters2016} to create a second-order minimum energy filter for the 15-DOF collaborative localisation problem.

The second-order minimum energy estimate, $\hat{X}$, for the state of the joint network of vehicles described above is
\begin{align}
\dot{\hat{X}} &= \hat{X} \left( \lambda(\hat{X}, u) + K r \right),
\label{eq:abstract_filter}
\\
\hat{X}(t_0) &= \argmin_X J_0(X),
\end{align}
where $Kr$ is a correction term composed of the gain operator $K$ acting on the residual $r$.
The residual, $r = r(\hat{X}, y) \in \uav(\nvehicles)^*$, is given by
\begin{multline}
r \coloneqq \sum_{\mathclap{i\in \bm{L}, \alpha \in \bm{V}}} T_eL^*_{\hat{X}} \circ \left[  \DQD^\alpha_i \circ (y^\alpha_i - \hat{y}^\alpha_i) \circ \der \hat{y}^\alpha_i \right]
\\
+  \sum_{\mathclap{\alpha, \beta \in \bm{V}}} T_eL^*_{\hat{X}} \circ \left[  \DQD^\alpha_\beta \circ (y^\alpha_\beta - \hat{y}^\alpha_\beta) \circ \der \hat{y}_\beta^\alpha \right]
\label{eq:def_r}
\end{multline}
where
\begin{align*}
    \hat{y}^\alpha_i &\coloneqq h(\hat{X}^\alpha, l_i),
    &
    \DQD^\alpha_i &\coloneqq  (D\inv)^\top Q^\alpha_i D\inv,
\\
    \hat{y}_\beta^\alpha &\coloneqq h(\hat{X}^\alpha, \hat{X}^\beta),
    &
    \DQD^\alpha_\beta &\coloneqq  (D\inv)^\top Q^\alpha_\beta D\inv.
\end{align*}
The gain operator, $K(t) : \uav(n)^* \rightarrow \uav(n)$, satisfies
\begin{multline}
    \dot{K} = A \circ K + K \circ A^* - K \circ E \circ K + B \circ W\inv \circ  B^*
    \\ - \connection_{Kr} \circ K - K \circ \connection^*_{Kr}
    \label{eq:K_dot}
\end{multline}
where $\Gamma$ is the connection function as in \eqref{eq:def_connection}. The initial condition is 
\begin{align*}
K(t_0) &= (T_eL^*_{\hat{X}_0} \circ \Hess J_0(\hat{X}_0) \circ T_eL_{\hat{X}_0})\inv,
\end{align*}
and the operators $A= A(\hat{X}, u)$ and $E = E(\hat{X}, y)$ are defined by
\begin{align}
A &\coloneqq \der_1 \lambda(\hat{X}, u) \circ T_eL_{\hat{X}} - \ad_{\lambda(\hat{X}, u)} - T_{\lambda(\hat{X}, u)},
\label{eq:def_A}
\\
E &\coloneqq  - T_eL^*_{\hat{X}} \circ  \left[
    \quad \sum_{\mathclap{ i\in \bm{L}, \alpha \in \bm{V}}} E^\alpha_i
    +  \sum_{\mathclap{\alpha, \beta \in \bm{V}}} E^\alpha_\beta
    \right] \circ T_eL_{\hat{X}},
    \label{eq:def_E}
\end{align}
\begin{gather*}
    \scalebox{0.93}{$
    E^\alpha_i \coloneqq  \bigl( \DQD^\alpha_i \circ (y^\alpha_i - \hat{y}^\alpha_i) \bigr) ^ {T_{\hat{X}}G} \circ \Hess \hat{y}^\alpha_i
    - (\der \hat{y}^\alpha_i)^* \circ \DQD^\alpha_i  \circ \der  \hat{y}^\alpha_i
    $},
    \\
    \scalebox{0.93}{$
    E^\alpha_\beta \coloneqq  \bigl( \DQD^\alpha_\beta \circ (y^\alpha_\beta - \hat{y}^\alpha_\beta) \bigr) ^ {T_{\hat{X}}G} \circ \Hess \hat{y}^\alpha_\beta
    - (\der \hat{y}^\alpha_\beta)^* \circ \DQD^\alpha_\beta  \circ \der  \hat{y}^\alpha_\beta
    $}.
\end{gather*}
% \begin{multline}
% E(t) =  - T_eL^*_{\hat{X}} \circ  \Big[
%     \\ \quad \sum_{\mathclap{ i\in \bm{L}, \alpha \in \bm{V}}}  \left( \DQD^\alpha_i \circ (y^\alpha_i - \hat{y}^\alpha_i) \right) ^ {T_{\hat{X}}G} \circ \Hess \hat{y}^\alpha_i
% - (\der \hat{y}^\alpha_i)^* \circ \DQD^\alpha_i  \circ \der  \hat{y}^\alpha_i)
% \\
% + \sum_{\mathclap{\alpha, \beta \in \bm{V}}}  \left( \DQD^\alpha_\beta \circ (y^\alpha_\beta - \hat{y}^\alpha_\beta) \right) ^ {T_{\hat{X}}G} \circ \Hess \hat{y}^\alpha_\beta
% - (\der \hat{y}^\alpha_\beta)^* \circ \DQD^\alpha_\beta  \circ \der  \hat{y}^\alpha_\beta)
% \\ \Big] \circ T_eL_{\hat{X}}.
% \end{multline}

\subsection{Explicit Matrix Representation}
% For example, the operator $A : \uav(\nvehicles) \rightarrow \uav(\nvehicles)$ can be equivalently be represented as a matrix operator, $\check{A} : \Real{15 \nvehicles} \rightarrow \Real{15 \nvehicles}$ such that
% \begin{align*}
%     A \circ \Psi = (\check{A} \Psi^\vee) ^\wedge
% \end{align*}
% for $\Psi \in \uav(\nvehicles)$. This allows us to represent \eqref{eq:K_dot} as a matrix ODE which can then be implemented numerically in software.

The filter equations shown in Section \ref{sec:minimum_energy_abstract}, describe a set of abstract operators on the Lie group $\UAV(\nvehicles)$ and the corresponding Lie algebra.
In this section we present a matrix representation of the filter, which enables implementation on a computer.

The filter equation \eqref{eq:abstract_filter} can be equivalently represented by 
\begin{gather}
    \dot{\hat{X}} = \hat{X} \left( \lambda^\vee(\hat{X}, u) + \check{K}\check{r}\right)^\wedge
    \label{eq:x_hat_matrix}
\end{gather}
where $\check{K} \in \Real{15 \nvehicles \times 15 \nvehicles}$ is the matrix representation of $K$.
The vector representation, $\check{r} \in \Real{15\nvehicles} $, of \eqref{eq:def_r} is
\begin{gather*}
    \check{r} \coloneqq \sum_{\mathclap{ i\in \bm{L}, \alpha \in \bm{V}}} (s^\alpha_i)^\top  F^\alpha_i(\hat{X})
    + \sum_{\mathclap{\alpha, \beta \in \bm{V}}} (s^\alpha_\beta)^\top F^\alpha_\beta(\hat{X})
 \end{gather*}
 where
 \begin{align*}
     s^\alpha_i &\coloneqq \DQD^\alpha_i (y^\alpha_i - \hat{y}^\alpha_i),
     \\
     s^\alpha_\beta &\coloneqq \DQD^\alpha_\beta (y^\alpha_\beta - \hat{y}^\alpha_\beta),
     \\
     F^\alpha_i(X) &\coloneqq \begin{bmatrix} \bar{F}_1 & \cdots & \bar{F}_{n} \end{bmatrix},
     \\
     \bar{F}_\alpha &\coloneqq \begin{bmatrix} h(X^\alpha, l_i)_\times & -\bm{I}_3 & \bm{0}_{3\times 9} \end{bmatrix},
     \\
     \bar{F}_j &\coloneqq \bm{0}_{3 \times 15} ~ (j \neq \alpha),
     \\
     F^\alpha_\beta(X) &\coloneqq  \begin{bmatrix} \mathring{F}_1 & \cdots & \mathring{F}_{n}\end{bmatrix},
     \\
     \mathring{F}_\alpha &\coloneqq \begin{bmatrix} h(X^\alpha, X^\beta)_\times & -\bm{I}_3 & \bm{0}_{3\times 9}\end{bmatrix},
     \\
     \mathring{F}_\beta  &\coloneqq \begin{bmatrix} - R_{\alpha \beta} m_{\beta \times} & R_{\alpha \beta} & \bm{0}_{3\times 9} \end{bmatrix},
     \\
     \mathring{F}_j  &\coloneqq \bm{0}_{3 \times 15} ~ (j \notin \{\alpha, \beta\}).
 \end{align*}

 Equivalently to \eqref{eq:K_dot}, the gain matrix $\check{K}$ satisfies
\begin{multline}
    \dot{\check{K}} = \check{A} \check{K} + \check{K} \check{A}^\top - \check{K} \check{E} \check{K} + \check{B} \check{W}\inv \check{B}^\top
    \\
    - \frac{1}{2} \check{\ad}_{K r} \check{K} - \frac{1}{2} \check{K} (\check{\ad}_{K r})^\top,
    \label{eq:K_matrix}
\end{multline}
where $\check{B}$ is from \eqref{eq:def_check_B} and $\check{A} \in \Real{15 \nvehicles \times 15 \nvehicles}$ is the matrix representation of \eqref{eq:def_A}, given by
\begingroup
\setlength{\arraycolsep}{4pt} % Default value: 6pt
\begin{gather*}
\check{A} \coloneqq \blkdiag(\check{A}^1, \ldots, \check{A}^{\nvehicles}),
\\
\check{A}^\alpha \coloneqq-\begin{bmatrix}
   (u^\alpha_\omega - \hat{\wbias}^\alpha)_\times & \bm{0} & \bm{0} & \bm{I} & \bm{0} \\
   \bm{0} &  (u^\alpha_\omega - \hat{\wbias}^\alpha)_\times & - \bm{I} & \bm{0} & \bm{0} \\
   (u^\alpha_a - \hat{\abias}^\alpha)_\times &  \bm{0} &(u^\alpha_\omega - \hat{\wbias}^\alpha)_\times & \bm{0} & \bm{I} \\
   \bm{0} & \bm{0} & \bm{0} & \bm{0} & \bm{0}  \\
   \bm{0} & \bm{0} & \bm{0} & \bm{0} & \bm{0} 
\end{bmatrix}
\end{gather*}
\endgroup
The matrix representation of \eqref{eq:def_E}, $\check{E} \in \Real{15 \nvehicles \times 15\nvehicles}$, is
\begin{gather*}
    \check{E} \coloneqq \sum_{\mathclap{ i\in \bm{L}, \alpha \in \bm{V}}} \check{E}^\alpha_i
    + \sum_{\mathclap{\alpha, \beta \in \bm{V}}} \check{E}^\alpha_\beta,
    \\
\begin{multlined}
\check{E}^\alpha_i \coloneqq 
    \projsym{ G(s^\alpha_i)^\top F^\alpha_i(X)} 
    + F^\alpha_i(X)^\top \DQD^\alpha_i F^\alpha_i(X),
\\
\shoveleft \check{E}^\alpha_\beta \coloneqq
    \mathbb{P}_s \Big(
    G_\alpha(s^\alpha_\beta)^\top F^\alpha_\beta(X)
    + G_\alpha(s^\alpha_\beta)^\top  R_{\alpha \beta} L^\alpha_\beta
    \\
    - G_\beta({\Rab}^\top s^\alpha_\beta)^\top L^\alpha_\beta \Big)
   + F^\alpha_\beta(X)^\top \DQD^\alpha_\beta F^\alpha_\beta(X) ,
\end{multlined}
\end{gather*}
where
\begin{align*}
G_\alpha(s) &\coloneqq \begin{bmatrix} \bar{G}_1 & \ldots & \bar{G}_n \end{bmatrix},
\\
\bar{G}_\alpha &\coloneqq \begin{bmatrix} s_\times & \bm{0}_{3\times 12} \end{bmatrix},
&
\bar{G}_j  &\coloneqq \bm{0}_{3 \times 15}~ (j\neq \alpha),
\\
L_\beta &\coloneqq \begin{bmatrix} \bar{L}_1 & \ldots & \bar{L}_n \end{bmatrix},
\\
\bar{L}_\beta &\coloneqq \begin{bmatrix} -m_{\beta\times} & \bm{I}_3 & \bm{0}_{3\times9} \end{bmatrix},
&
\bar{L}_j &\coloneqq \bm{0}_{3\times 15}~(j \neq \beta) ,
\\
\Rab &\coloneqq (R^\alpha)^\top R^\beta.
\end{align*}
\section{A Discrete-Time Filter Implementation}
\label{sec:discrete_time}
The filter proposed in Section \ref{sec:joint_filter} considers the system in continuous-time, with all measurements available at all times.
However, due to the discrete nature of digital sensors and computers a discrete-time implementation is necessary.
Modern IMU sensors are capable of sampling rates in the hundreds of Hz, and some into the kHz range.
On the other hand, a camera based system of detecting landmarks and other vehicles may only be capable of sampling rates in the order of 10 to 100 Hz.

Additionally, while it is reasonable to assume that IMU measurements are uninterrupted, it is unlikely that all landmark and inter-vehicle measurements will be available at all sampling times.
A number of factors, including the range and field-of-view of the sensor, occlusion, and interference, means that landmark and inter-vehicle measurements may only be available sporadically and with variable intervals between measurements.

Our proposed method of discretisation relies on numerically integrating the terms in \eqref{eq:x_hat_matrix} and \eqref{eq:K_matrix} which correspond to the external measurements separately to the IMU measurements and at different time intervals.
If we firstly just consider the terms related to the IMU measurement, we have
\begin{align}
    \dot{\hat{X}} &= \hat{X} \lambda(\hat{X}, u),
    \label{eq:x_hat_dot_imu}
    \\
    \dot{\check{K}} &= \check{A} \check{K} + \check{K} \check{A}^\top + \check{B} \check{W}\inv \check{B}^\top.
    \label{eq:K_imu_continuous}
\end{align}
We assume that the IMU has a fixed sampling rate of $f_u$ Hz, with a corresponding sampling period of $\Delta t_u$ seconds, and will use a sample-and-hold strategy.
It is then straightforward to numerically integrate \eqref{eq:x_hat_dot_imu} and \eqref{eq:K_imu_continuous} forward in time from a time $t$ to time $t+ \Delta t_u$,
\begin{align}
    \hat{X}(t+\Delta t_u) &= \hat{X}(t) \exp(\Delta t_u \cdot \lambda(\hat{X}, u)),
    \label{eq:X_hat_imu_discrete}
    \\
    \check{K}(t+ \Delta t_u) &= \check{K}(t) + \Delta t_u (\check{A} \check{K}(t) + \check{K}(t) \check{A}^\top + \check{B} \check{W}\inv \check{B}^\top) \nonumber % \label{eq:K_imu_discrete}
\end{align}
where $\exp$ is the exponential operator on $\UAV(n)$.

We apply the same strategy to the external measurements, but must make some additional considerations.
Considering a single landmark measurement, $y^\alpha_i$, the relevant terms in the state estimate ODEs are
\begin{align*}
    \dot{\hat{X}} &= \hat{X} \left(\check{K}(t) \check{r}^\alpha_i(\hat{X})\right)^\wedge,
    \\
    \dot{\check{K}} &=  - \check{K} \check{E}^\alpha_i \check{K} -\projsym{\check{\ad}_{K r^\alpha_i} \check{K}}.
\end{align*}

While the frequency of measurements may not be constant, we can still measure the period between subsequent landmark measurements, which we denote as $\Delta t^\alpha_i$. 
A landmark measurement received at time $t$ can be numerically integrated in a similar way as the IMU measurements;
\begin{align}
    \hat{X}(t^+) &= \hat{X}(t) \cdot \exp(\Delta t^\alpha_i \cdot \check{K}(t^+) \check{r}^\alpha_i(\hat{X}))
    \label{eq:centralised_landmark_update_estimate}
    \\
    K(t^+) &= \left(I + \Delta t^\alpha_i \check{K}(t) ( \check{E}^\alpha_i + \projsym{\check{K}\inv(t) \check{\ad}_{K r^\alpha_i} }) \right)\inv \check{K}(t)
    \label{eq:centralised_landmark_update_K}
\end{align}

In practice, $t$ must be a multiple of $\Delta t_u$, and so the precise time the landmark measurement is received is rounded up to the next multiple of $\Delta t_u$.
Given that $\Delta t_u$ is small and considering the velocities of vehicles within this time-frame, this approximation introduces negligible error.
Measurements of different landmarks which are received within the same time interval are processed sequentially and it is straightforward to see how the same process can be used to integrate inter-vehicle measurements.

\subsection{Tuning}
A practical consideration to make when implementing any filter is the choice of different tuning parameters.
For the minimum-energy filter presented in this paper, there is free choice in $B, D, W, Q$, and the cost, $J_0$, on the initial estimate.
$J_0$ is typically implicitly defined by initialising the filter with a-priori information about the initial state of the system.
When choosing the remaining parameters, it is necessary to have an understanding of the behaviour of the physical sensors in the system.
A convenient choice for $B$ and $D$ is the square root of the covariance of the sensor measurement error, which can be measured and calibrated a priori.

In choosing the sensor covariance, we must also account for the difference between the continuous time model used in the cost functional, and the discrete-time implementation of the filter equations.
The sample-and-hold strategy means that the measurement signals are correlated almost perfectly for the length of the hold time.
This poses an issue when the sample rate of the IMU sensor is different to that of the external measurements.
Consequently, we set $W= \frac{1}{\Delta t_u} \bm{I}$ and $Q^\alpha_i=\frac{1}{\Delta t^\alpha_i} \bm{I}$ to appropriately weight the terms in the cost functional and correct for the differences in sample rates.

As with any filter, the suggested values above provide an initial starting point for tuning.
It may be necessary to adjust these values, particularly $B$, to maintain filter stability and robustness.
\section{Distributing the Multi Vehicle Filter}
\label{sec:distributed-filter}

The filter derived in Sections \ref{sec:joint_filter} and \ref{sec:discrete_time} is a single set of equations which simultaneously estimates the state of all vehicles in the network.
Implementing these equations in a physical system would require a central processing node to collect the measurements from every vehicle in the network and provide state estimates of each vehicle.
This not only introduces a single point of failure in the network but also imposes significant communication overheads as all IMU data and measurement data has to be transmitted to a central location.

Similar to our approach in \cite{hendersonMinimumEnergyFilter2020IFAC}, we show how the centralised filter equations can be decoupled so that each vehicle stores its own state estimate, and communication is limited to times when external measurements are received.
For each vehicle, $\alpha \in \bm{V}$, an onboard processing node will store the vehicle's own state estimate, $\hat{X}^\alpha$, as well as $\check{K}^\alpha \in \Real{15 \nvehicles \times 15}$, the sub-matrix of $\check{K}$ corresponding to the vehicle $\alpha$.

By observation, it is trivial to decouple the IMU propagation step from \eqref{eq:X_hat_imu_discrete}, which is
\begin{align}
    \hat{X}^\alpha(t+ \Delta t_u) &= \hat{X}^\alpha(t) \cdot \exp( \Delta t_u \cdot \lambda^\alpha(\hat{X}^\alpha, u^\alpha)).
    \label{eq:distributed_state_propagation}
\end{align}

By utilising the fact that $\check{B}$ and $\check{W}$ are block diagonal and $\check{K}$ is symmetric, \eqref{eq:K_imu_continuous} can be partitioned into a set of $\nvehicles \times \nvehicles$ block matrices, each of size $15 \times 15$. The diagonal blocks of $\check{K}$ satisfy
\begin{align}
    \dot{\check{K}}^\alpha_\alpha &= \check{A}^\alpha \check{K}^\alpha_\alpha + \check{K}^\alpha_\alpha (\check{A}^\alpha)^\top + \check{B}^\alpha (\check{W}^\alpha)\inv (\check{B}^\alpha)^\top,
    \label{eq:K_aa_distributed}
\end{align}
and the off-diagonal blocks, $\check{K}^\alpha_\beta$, satisfy
\begin{align}
    \dot{\check{K}}^\alpha_\beta &= \check{A}^\alpha \check{K}^\alpha_\beta + \check{K}^\alpha_\beta (\check{A}^\beta)^\top.
    \label{eq:K_ab_distributed}
\end{align}

Given that we apply a sample-and-hold strategy to the IMU measurements, the $\check{A}$ matrices are constant for a period of $\Delta t_u$. This results in an explicit solution to \eqref{eq:K_ab_distributed};
\begin{align*}
    \check{K}^\alpha_\beta(t+\Delta t_u) &= \exp(\check{A}^\alpha(t) \Delta t_u ) \check{K}^\alpha_\beta(t) \exp((\check{A}^\beta(t))^\top \Delta t_u ).
\end{align*}
Multiple consecutive IMU measurements can be incorporated by repeatedly applying the equation above. For $k$ consecutive measurements this gives
\begin{gather}
    \check{K}^\alpha_\beta(t + k \Delta t_u) = \Lambda^\alpha \check{K}^\alpha_\beta(t) (\Lambda^\beta)^\top,
    \label{eq:K_ab_factored}
    \\
    \Lambda^\alpha \coloneqq \prod_{j=k}^1 \exp\left(\check{A}^\alpha(t + j \Delta t_u) \Delta t_u \right).
\end{gather}
The key property of \eqref{eq:K_ab_factored} is that $\Lambda_\alpha$ and $\Lambda_\beta$ can be computed independently on vehicles $\alpha$ and $\beta$, respectively.

Assuming no external measurements are made, each vehicle can independently propagate forward in time its own state estimate from \eqref{eq:distributed_state_propagation}, and the diagonal element of $K$ by discretising \eqref{eq:K_aa_distributed}. Each vehicle also computes and stores $\Lambda_\alpha$ and, at some point in time, it can communicate with another vehicle, $\beta$, to receive $\Lambda_\beta$ and then calculate $\check{K}^\alpha_\beta$. This means that in the time between external measurements, no communication between vehicles is required.

When an external measurement is made, for example a landmark measurement $y^\alpha_i$, vehicles must communicate with each other to share their $\Lambda$ terms and to receive update information. Equation \eqref{eq:centralised_landmark_update_estimate} can be easily decoupled, giving
\begin{align*}
\hat{X}^j(t^+) &= \hat{X}^j(t) \cdot \exp(\Delta t^\alpha_i \cdot \check{K}^{j \top}(t^+) \check{r}^\alpha_i) \quad\forall j \in \bm{V},
\end{align*}
while $\check{r}^\alpha_i$ can be computed by vehicle $\alpha$ and then communicated to others.

The update for $K$ presents more of a challenge due to the appearance of $K\inv$ in the curvature correction term of \eqref{eq:centralised_landmark_update_K}, $\projsym{\check{K}\inv \check{\ad}_{K r^\alpha_i} }$.
Computing the curvature correction term would require a centralised computation to calculate $K\inv$ and would negate the advantages of the decentralised filter implementation.
Instead, we  disregard the term and approximate \eqref{eq:centralised_landmark_update_K} as
\begin{align}
    K^j(t^+) &\approx \left(I + \Delta t^\alpha_i \check{K}(t) \check{E}^\alpha_i(t) \right)\inv \check{K}^j(t).
    \label{eq:decentralised_landmark_update_K}
\end{align}
In Section \ref{sec:simulation}, we compare simulation results from the centralised filter to the decentralised version and can observe that removing the curvature term in the decentralised implementation has only a minor effect on the performance of the filter, primarily in the transient region. The term $\check{E}^\alpha_i$ can be calculated locally on vehicle $\alpha$, and, due to the sparsity properties of this term, it is possible to calculate $\check{K}\check{E}^\alpha_i$ with only the terms in $\check{K}^\alpha$. Thus, the entire term to be inverted in \eqref{eq:decentralised_landmark_update_K} can be calculated locally on vehicle $\alpha$ and then shared with every other vehicle $j$ to update $\check{K}^j$.

In summary, the decentralised filter allows vehicles to operate completely independently during periods where no external measurements are received. At a time when an external measurement is received by a vehicle, all other vehicles must communicate their computed $\Lambda$ terms to each other. The vehicle, $\alpha$, which makes the measurement computes $r^\alpha_i$ and the first term in \eqref{eq:decentralised_landmark_update_K}, and transmits this information to the other vehicles in the network. On receiving this information, the other vehicles update their own values of $\hat{X}^j$ and $\check{K}^j$.

The update process for inter-vehicle measurements follows in the same fashion, however in order to calculate the terms $\check{r}^\alpha_\beta$ and $\check{E}^\alpha_\beta$, the terms $\hat{X}^\beta$ and $\check{K}^\beta$ must additionally be communicated from vehicle $\beta$ to $\alpha$.

\section{Simulation}
\label{sec:simulation}

In this section, we demonstrate the application of the proposed multi-vehicle filter through a simulation with a mix of real-world and synthetic data.
The EuRoC MAV dataset \cite{burriEuRoCMicroAerial2016} is a sequence of 11 trials of a single micro aerial vehicle (MAV) flying in an indoor space and is primarily intended for benchmarking Simultaneous Localisation and Mapping (SLAM) algorithms.
Each trial records IMU data from the MAV, images from onboard stereo cameras, and a ground truth position of the MAV w.r.t. a global reference frame which is captured from a VICON motion capture system.

The EuRoC MAV dataset provides us with realistic flight trajectories as well as real-world IMU data to evaluate our algorithm on.
We discard the stereo camera data, and instead generate synthetic landmark measurements from 3 virtual landmarks placed at fixed locations.
To create a collaborative localisation scenario, we merge a number of different single-vehicle trials into one combined dataset, which simulates multiple vehicles operating concurrently in the same environment.
We then also generate a set of synthetic relative position measurements between vehicles.

\begin{figure}
    \includegraphics[width=\linewidth]{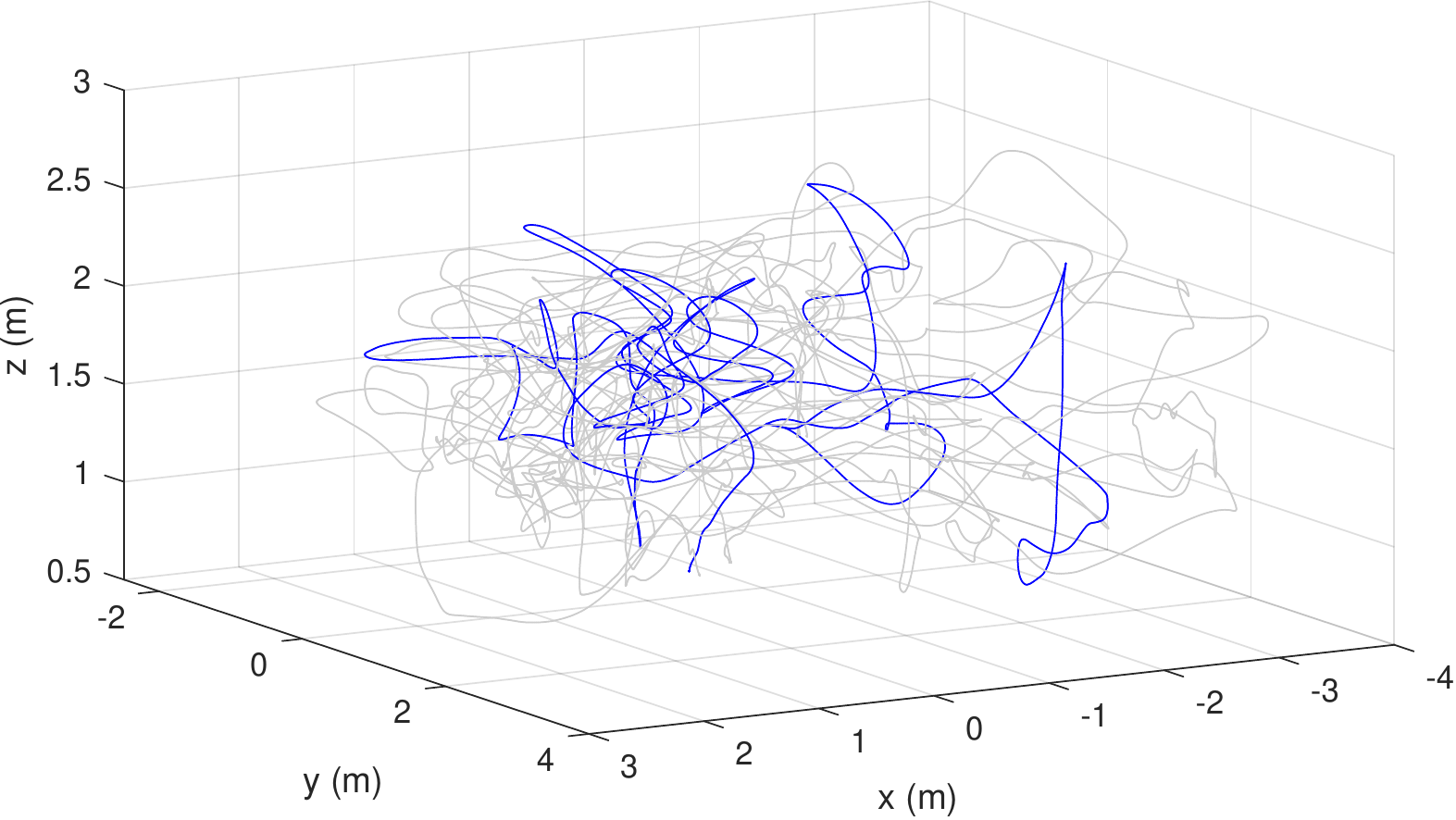}
    \caption{Trajectories of the 6 vehicles in the simulation based on real-world UAV flights from the EuRoC dataset \cite{burriEuRoCMicroAerial2016}. The trajectory of a single vehicle has been highlighted in blue}
    \label{fig:trajectory}
\end{figure}

We combine the 6 trials from the EuRoC dataset which were captured in the VICON room (V101, V102, V103, V201, V202, V203) into a multi-robot dataset.
The length of the dataset is 83.5 seconds, which is the maximum length for which there is data from all trials.
Figure \ref{fig:trajectory} shows the trajectories of each of the vehicles in the combined dataset.
Real-world IMU data for each vehicle is available at a rate of 200 Hz.
Synthetic landmark measurements to the 3 virtual landmarks in the room are available at a rate of 10 Hz, and inter-vehicle measurements between all vehicles are available at a rate of 10 Hz.
The initial estimate for each vehicle is initialised with an approximate pose of the vehicle and with zero velocity and zero bias.

\begin{figure}
    \includegraphics[width=0.95\linewidth]{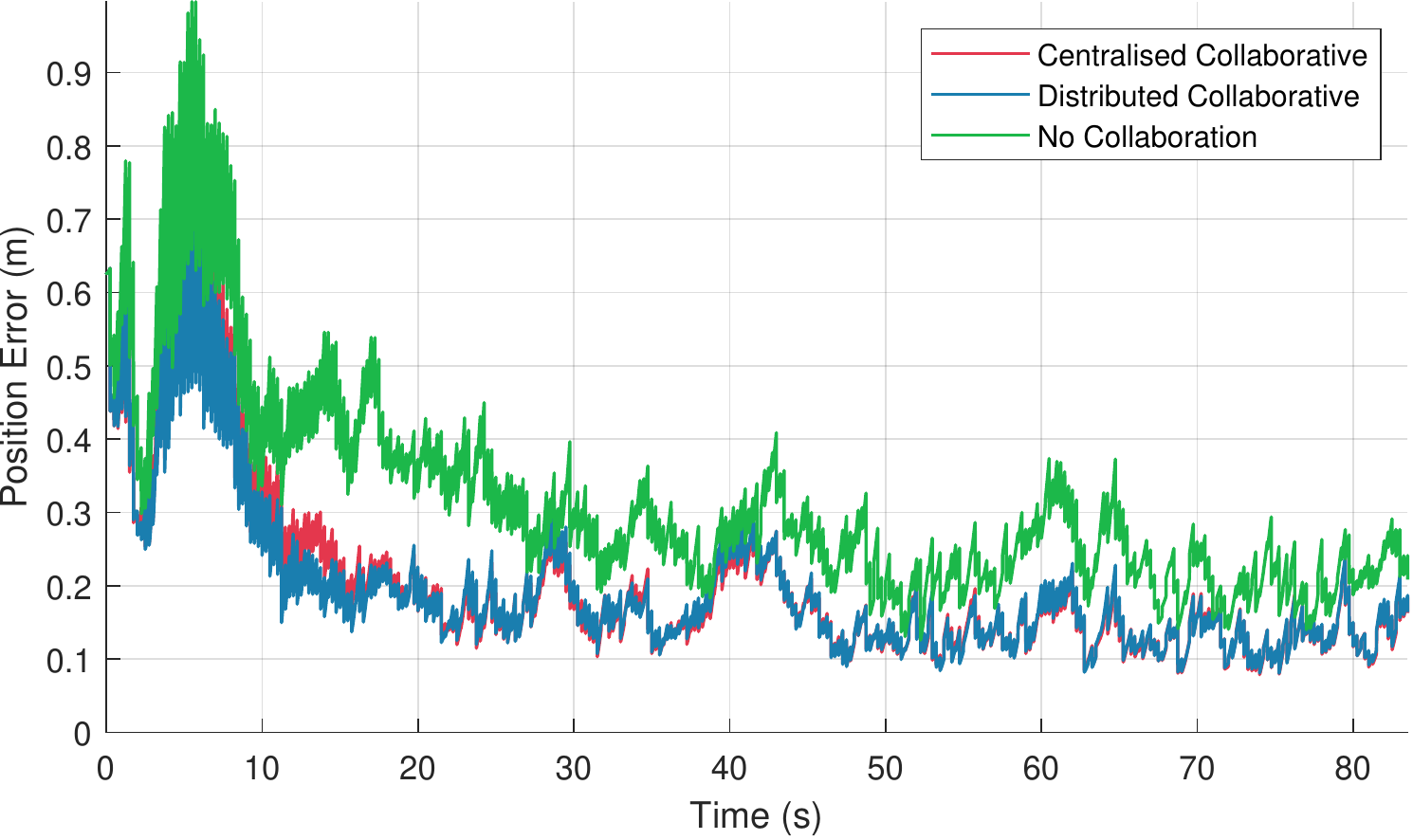}
    \caption{Position error in filter estimate over time comparing the collaborative location filter with the non-collaborative filter. Values shown are the mean errors for the 6 vehicles in the dataset.}
    \label{fig:position_error}
\end{figure}

\begin{figure}
    \includegraphics[width=0.95\linewidth]{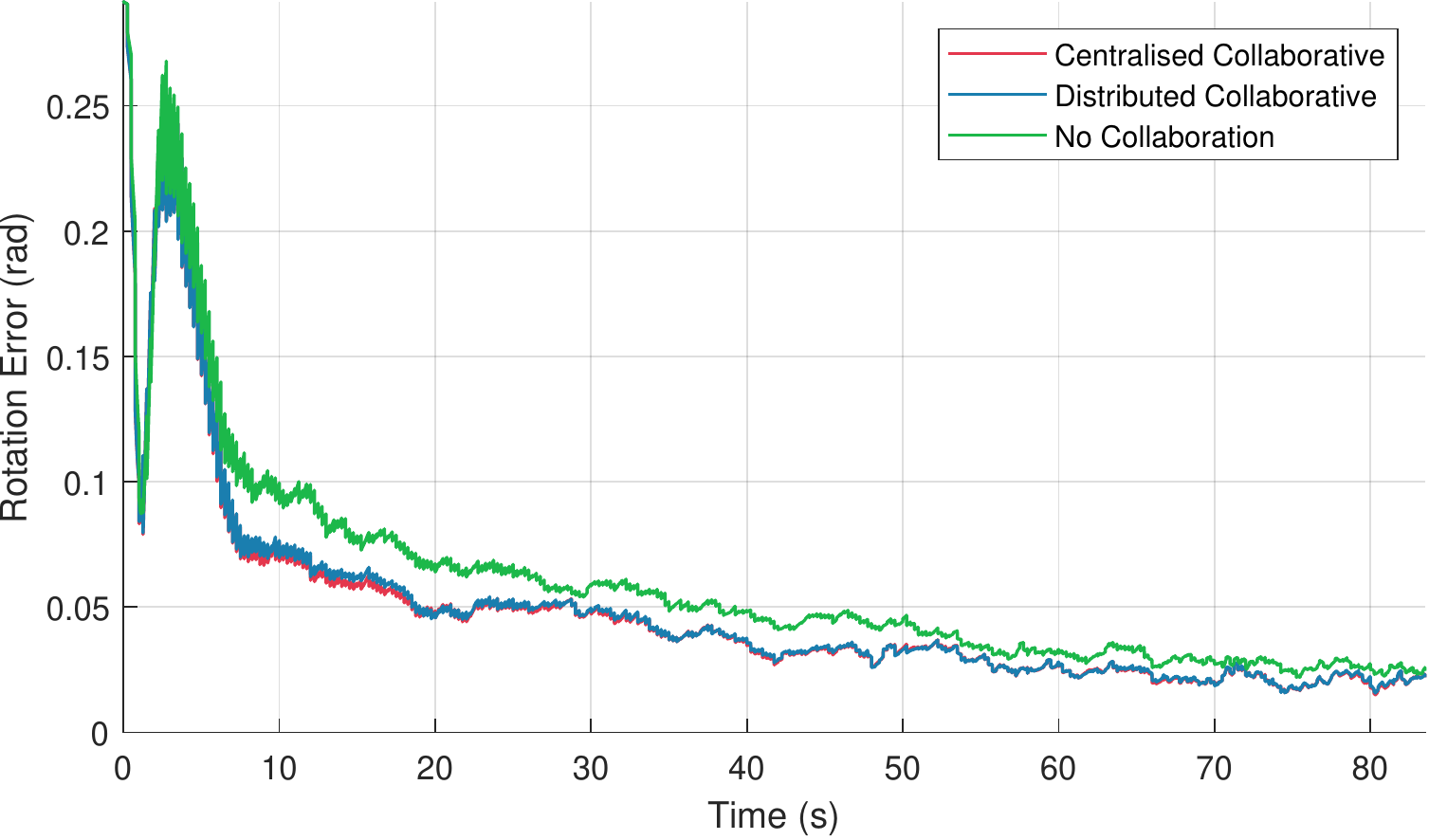}
    \caption{Rotation error in filter estimate in simulation. Values shown are the mean errors for the 6 vehicles in the dataset.}
    \label{fig:rotation_error}
\end{figure}

We simulate three different versions of the filter.
Firstly, as a baseline comparison, we simulate a non-collaborative filter where each vehicle operates independently and does not utilise the inter-vehicle measurements.
We also simulate the centralised collaborative filter \eqref{eq:X_hat_imu_discrete}-\eqref{eq:centralised_landmark_update_K} which includes the curvature correction terms, and finally we simulate the distributed collaborative filter \eqref{eq:distributed_state_propagation}-\eqref{eq:decentralised_landmark_update_K}.
Figure \ref{fig:position_error} shows the error between the ground truth position and the filter estimate of position, measured as the Euclidean distance and averaged over the 6 vehicles.
Figure \ref{fig:rotation_error} shows the error in the rotation estimate, measured as $|\tilde{\theta}| = \arccos(\frac{\tr(\hat{X}_R\inv X_R) -1}{2})$, also averaged over the 6 vehicles.
We present a summary of the simulation results in Table \ref{tab:simulation_error}.

\begin{table}
    \caption{Average estimation error over the total simulation time}
    \begin{center}
    \begin{tabular}{l|c|c|c}
                                & No Collab.  & Centralised  & Decentralised \\
        \hline
        Position (m)                 & 0.315             & 0.202                 & \textbf{0.197}             \\
        Rotation (rad)               & 0.059             & \textbf{0.048}                 & 0.049             \\
        Linear Velocity (m/s)        & 0.340             & \textbf{0.254}                 & 0.256             \\
        IMU Gyro Bias  (rad/s)       & 0.021             & 0.021                 & 0.021             \\
        IMU Accel. Bias  (m/s$^2$)   & 0.206             & \textbf{0.167}                 & 0.190
    \end{tabular}
    \label{tab:simulation_error}
\end{center}
\end{table}

There are a number of different simulations we could have presented to show scenarios where non-collaborative localisation will rapidly diverge, for example if some vehicles have no access to landmark measurements. In these cases, collaborative localisation is clearly superior as vehicles can help localise each other through inter-vehicle measurements. However, this simulation shows that, even when each individual vehicle has an abundance of landmark measurements, collaborative localisation can further improve the accuracy of the state estimate.

A number of interesting features arise from the simulation results.
Primarily, the collaborative filters clearly outperform the non-collaborative filter.
At every point in time, both versions of the collaborative filter have a lower position error than the non-collaborative filter.
And for more than 99\% of the time, both collaborative filters outperform the rotation estimation of the non-collaborative filter.
On average, the position error of the distributed collaborative filter is 37\% lower and the rotation error is 17\% lower compared to the non-collaborative filter.

In the first 10 seconds of the simulation, there is both a high error as well as a high variability in the error for all filters.
This is because the estimate of bias in the IMU is initialised to zero and must converge to the true value.
While the errors in the bias estimates are high, the double integration of acceleration measurements will result in highly inaccurate position estimates.
As more sensor information is collected, the bias can be accurately estimated and the variation in error decreases significantly.

The difference in performance between the centralised and decentralised collaborative filters is a result of not being able to distribute the curvature correction term in \eqref{eq:centralised_landmark_update_K}.
It appears that the effect of the curvature term is greatest during the transient phase, but once the filter converges, the difference is negligible.
Further research is required to determine whether this term has a significant effect in some scenarios, and whether it is possible to also decouple the computation so that it can be incorporated into the distributed filter.

\section{Conclusion}
\label{sec:conclusion}
In this paper we have have presented a distributed, inertial, collaborative localisation algorithm based on the principle of minium energy filtering.
We have demonstrated the performance of the filter in a simulation based on real-world IMU data, and it is capable of being implemented on a real-world system.
In our future work on this topic, we plan to compare our approach with other existing collaborative localisation algorithms and to relax further the communication requirements of the filter. We also plan to demonstrate the algorithm on a physical network of both unmanned aerial vehicles and unmanned ground vehicles.

\renewcommand*{\bibfont}{\footnotesize}

\printbibliography

\end{document}